\documentclass[twocolumn]{article}
\usepackage{arxiv}
\usepackage{amsmath}
\usepackage[utf8]{inputenc}
\usepackage[T1]{fontenc}
\usepackage{url}
\usepackage{booktabs}
\usepackage{amsfonts}
\usepackage{nicefrac}
\usepackage{microtype}
\usepackage{lipsum}
\usepackage{graphicx}
\usepackage{enumitem}
\usepackage{wrapfig}
\usepackage{multirow}
\usepackage{wasysym}
\usepackage{makecell}
\usepackage{pifont}
\usepackage{colortbl}
\usepackage{caption}
\usepackage[hidelinks]{hyperref}
\usepackage{tabularx}
\usepackage{tcolorbox}
\usepackage{orcidlink}
\usepackage{xcolor}
\usepackage{soul}
\usepackage{array}
\usepackage{calc}
\usepackage{float}
\usepackage{tikz}
\usetikzlibrary{shapes.misc}

\graphicspath{ {./images/} }

\definecolor{colorVU}{RGB}{220, 248, 220}
\definecolor{colorDD}{RGB}{255, 225, 195}
\definecolor{colorML}{RGB}{235, 248, 255}
\definecolor{colorQA}{RGB}{250, 235, 250}
\definecolor{colorMM}{RGB}{255, 255, 220}
\definecolor{colorAvg}{RGB}{255, 228, 230}

\newcommand{\mname}{EmambaIR}
\begin{document}

\twocolumn[{%
\begin{center}
\rule{\textwidth}{2.8pt}  

\vspace{1.1em}

{\Large\bfseries \textit{EmambaIR}: Efficient Visual State Space Model for Event-guided
Image Reconstruction}

\vspace{1.1em}
\rule{\textwidth}{1.0pt}  

\vspace{1.2em}  

{\normalsize  
\textbf{Wei Yu}$^{1\dagger}$ \quad
\textbf{Yunhang Qian}$^{1\dagger}$
}

\vspace{0.8em}

{\small
$^{1}$Harbin Institute of Technology \\
$^{\dagger}$Equal contribution
}

\end{center}
\vspace{2.5em} 
}] 

\centerline{\large\bfseries ABSTRACT}
\vspace{0.6em}

Recent event-based image reconstruction methods predominantly rely on Convolutional Neural Networks (CNNs) and Vision Transformers (ViTs) to process complementary event information. However, these architectures face fundamental limitations: CNNs often fail to capture global feature correlations, whereas ViTs incur quadratic computational complexity (e.g., $O(n^2)$), hindering their application in high-resolution scenarios. To address these bottlenecks, we introduce EmambaIR, an Efficient visual State Space Model designed for image reconstruction using spatially sparse and temporally continuous event streams. Our framework introduces two key components: the cross-modal Top-k Sparse Attention Module (TSAM) and the Gated State-Space Module (GSSM). TSAM efficiently performs pixel-level top-k sparse attention to guide cross-modal interactions, yielding rich yet sparse fusion features. Subsequently, GSSM utilizes a nonlinear gated unit to enhance the temporal representation of vanilla linear-complexity ($O(n)$) SSMs, effectively capturing global contextual dependencies without the typical computational overhead. Extensive experiments on six datasets across three diverse image reconstruction tasks—motion deblurring, deraining, and High Dynamic Range (HDR) enhancement—demonstrate that EmambaIR significantly outperforms state-of-the-art methods while offering substantial reductions in memory consumption and computational cost. The source code and data are publicly available at: \href{https://github.com/YunhangWickert/EmambaIR}{https://github.com/YunhangWickert/EmambaIR}



\section{Introduction}
\begin{figure}[t!]
    \begin{center}
    \includegraphics[width=\linewidth]{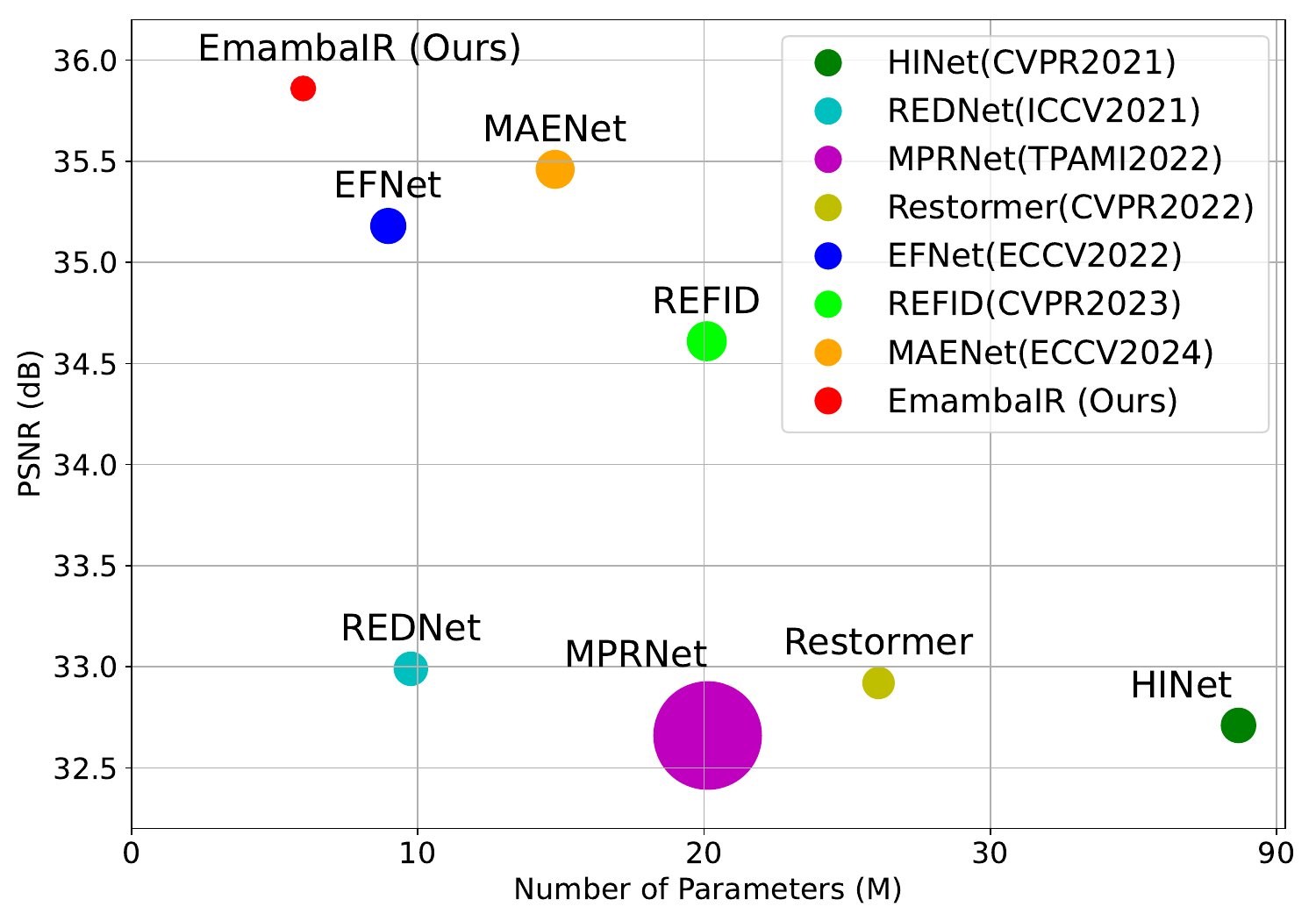}
    \end{center}
    \vspace{-0.2cm}
    \caption{
    A performance and efficiency comparison between existing CNNs-based and ViTs-based reconstruction methods and our SSM-based method shows that our \mname{} achieves superior performance with lower memory usage and computational cost.
    }
    \vspace{-0.3cm}
\label{fig:highight}
\end{figure}
Image-only based reconstruction algorithms~\cite{Blinddeconvolution, normalizedsparsity, Unnatural} have shown impressive progress, but they still struggle to restore degraded images under extreme scenes with high-speed motion and insufficient exposure. 
Recently, the bio-inspired event sensor has emerged as a powerful tool for image reconstruction. Unlike traditional cameras that capture entire image frames at fixed time intervals, event cameras asynchronously record per-pixel intensity changes (events) with microsecond-level temporal resolution and a high dynamic range (up to 120 dB). This fine granularity and extremely low latency make them exceptionally well-suited for addressing degradation problems caused by motion and lighting variations.

Benefiting from advancements in deep learning, various event-based image reconstruction methods have been developed to enhance image quality, including attention models~\cite{Spatially,BANet}, multi-scale fusion~\cite{GoPro,SRN}, multi-stage networks~\cite{chen2021hinet, MPRNet}, and coarse-to-fine strategies~\cite{Rethinking}.
These methods mainly adopt Convolutional Neural Networks (CNNs) and Visual Transformers (ViTs) to learn the fusion reconstruction of images and events. However, they face two primary limitations:
(i) CNN-based methods~\cite{Noise-trained, chen2023learning} primarily focus on local details and often overlook global contexts due to their intrinsic local receptive fields~\cite{pan2022integration}. This limitation hampers long-range feature aggregation, leading to reconstructed images that are visually unclear and more susceptible to noise and blur.
(ii) ViT-based methods~\cite{Eventbased, zamir2022restormer} alleviate such limitations by capturing non-local information, but their self-attention mechanism introduces a computational complexity that is quadratic ($\mathcal{O}(n^2)$) with respect to the input size $n$. This leads to high computational demands during both training and inference.
Consequently, these limitations restrict their practical effectiveness in high-resolution image reconstruction applications.

Recently, the State Space Model (SSM)~\cite{SSM} has garnered significant attention in Natural Language Processing (NLP) and high-level vision tasks~\cite{Visionmamba, Vlmamba, Segmamba} for its innovative, highly efficient network architecture. It showcases substantial advancements in long-range selection mechanisms and hardware efficiency optimization. Despite this potential, few studies have explored integrating event-based SSM methods to address the challenges of image reconstruction.
Inspired by this, we propose an efficient visual state space model for event-based image reconstruction tasks, namely EmambaIR, which is specifically designed to handle event streams characterized by spatial sparsity and temporal continuity.
It consists of cross-modal Top-k Sparse Attention Modules (TSAMs) and Gated State-Space Modules (GSSMs) to aggregate the complementary spatial and temporal features of events.
Specifically, to efficiently aggregate spatially sparse correlation features, our TSAM dynamically controls the sparsity and selectively guides the interaction of cross-modal features between events and images, obtaining sparse fusion features under the guidance of top-k sparse attention.
To reduce the computational complexity of high-resolution reconstruction, our GSSM employs a nonlinear gated unit to enhance the continuous temporal representation capabilities of the vanilla SSM. This allows the model to learn global gated features with long-range context correspondence while maintaining linear complexity ($\mathcal{O}(n)$).
Figure~\ref{fig:highight} shows the performance and efficiency comparison between existing state-of-the-art image motion deblurring reconstruction methods and our approach.
Our \mname{} achieves superior performance across various reconstruction tasks (e.g., deblurring, HDR, and deraining), while maintaining significant advantages in memory efficiency and computational cost.
Overall, our main contributions are summarized as follows:
\begin{itemize}
\item 
We propose a Top-k Sparse Attention Module (TSAM) that efficiently integrates the pixel-level features of events and images through dynamic sparsity selection, obtaining cross-modal fused features under the guidance of spatial top-k sparse attention.
\item 
We develop a Gated State Space Module (GSSM) to learn the channel-wise contextual correspondence of fused features. It employs nonlinear gated units to enhance the long-range representation capability of the vanilla SSM for continuous event streams.
\item 
Extensive experiments on synthetic and real datasets across three event-guided image reconstruction tasks demonstrate that our EmambaIR outperforms existing state-of-the-art approaches while requiring significantly lower computational costs.

\end{itemize}

\section{Related Work}

\subsection{Event-guided Image Reconstruction}
Leveraging the high dynamic range and microsecond-level temporal resolution of event streams, recent research has increasingly utilized events to guide high-quality image reconstruction.
Early work~\cite{eslnet} first introduced events to assist image reconstruction by proposing a sparse learning framework that performs end-to-end denoising and deblurring.
Since then, numerous approaches~\cite{Learning, kim2024frequency, sun2025motion} have developed advanced CNN- and ViT-based networks to integrate visual and temporal knowledge across both global and local scales, achieving more accurate and robust image deblurring.
Similarly, several methods~\cite{messikommer2022multi, yang2023learning, xiaopeng2024hdr} have introduced various event representation strategies and attention fusion modules to align signals with different dynamic ranges, enabling better HDR image restoration.
However, these reconstruction techniques typically rely on dense attention mechanisms to aggregate features without fully exploiting the inherent spatial sparsity of event streams. This limitation often leads to the suboptimal fusion of complementary information and unnecessarily high computational complexity.

\begin{figure*}[t!]
    \centering
    \includegraphics[width=\textwidth]{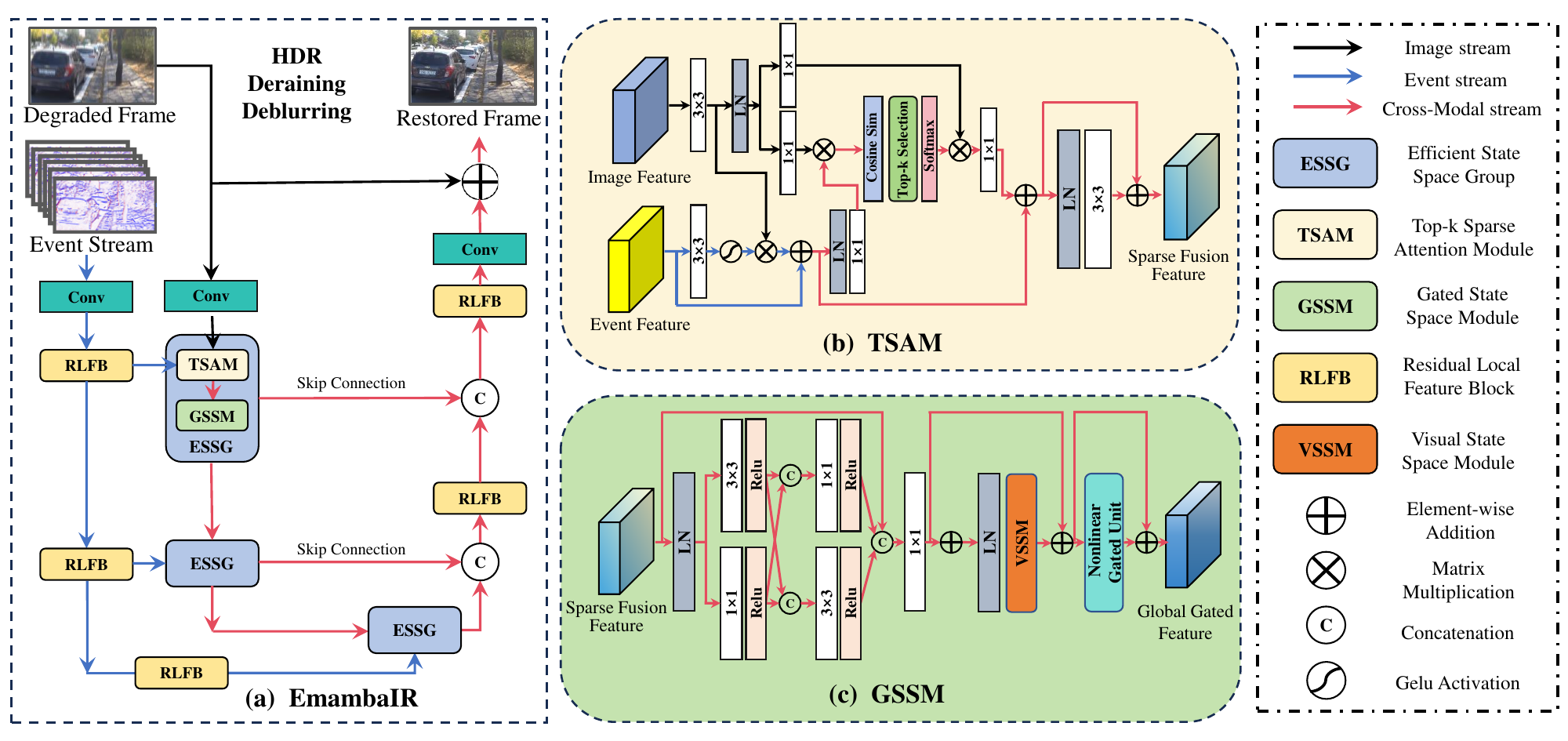}
    \vspace{-0.4cm}
    \caption{
    Overall architecture of (a) our EmambaIR for event-guided image reconstruction, which consists of a UNet-based backbone built upon the proposed (b) top-k sparse attention module and (c) gated state space module.
    }
\vspace{-0.5cm}
\label{fig:structure}
\end{figure*}

\subsection{Visual State Space Models}
Visual State Space Models (SSMs) have recently shown strong potential for long-range sequence modeling across various tasks.
Gu et al.~\cite{gu2021efficiently} and Islam et al.~\cite{islam2022long} proposed the Structured State-Space Sequence (S4) model as an efficient alternative to CNNs or Transformers for modeling long-range dependencies.
Building on this, recent studies~\cite{baron2023,ma2024u, Vmamba, Segmamba} have designed 2D SSMs to process visual data, supplanting traditional attention mechanisms with scalable SSM-based backbones to generate high-quality images.
In particular, Mamba improves upon S4 by introducing a selective mechanism and a hardware-aware efficient design.
Following this trend, several researchers~\cite{guo2025mambair, shi2024vmambair, he2024multi, gao2024learning} have explored the application of SSMs in low-level computer vision, often incorporating convolutions and channel attention to enhance the representation capabilities of standard Mamba architectures.
Despite these advances, the power of Mamba in capturing cross-modal interactions for event-guided image reconstruction remains largely unexplored.
\subsection{Sparse Representation}
%
%
Recent studies~\cite{Eventbased, eventdriven, tang2022sparse, chen2023learning} have investigated the sparse characteristics of event data, introducing local attention operations into CNN backbones. By restricting attention to local window sizes, these methods utilize sparse connection representations rather than full dense connections, thereby significantly reducing computational costs.
Because local attention modules only generate weights between adjacent elements, their computational complexity scales linearly with spatial resolution. Leveraging this, existing Transformer-based methods~\cite{zhang2022accurate, wang2022kvt, wang2022nformer, choi2024reciprocal} have introduced local inductive biases to enforce sparsity, allowing tokens from sparse areas to interact with global features efficiently. 
Unlike these existing representation methods, we implement a simple yet effective top-k sparse attention mechanism that approximates the global sparse properties of event streams, achieving highly efficient feature representation and cross-modal fusion.

\section{Proposed Method}

In Figure~\ref{fig:structure}, we present an Efficient visual State Space Model for Event-guided Image Reconstruction (EmambaIR) designed to restore degraded frame images using event streams. This architecture includes the commonly used Residual Local Feature Block (RLFB) alongside our proposed TSAMs and GSSMs, which are built upon the vanilla Visual State Space Module (VSSM).
Given a degraded image and its corresponding continuous event stream, we first extract the image and event features using standard convolutional layers. 
To achieve cross-modal information interaction, we feed both the image and event features into the TSAM to obtain sparse fusion features. This approach significantly reduces the inference time required for event feature interaction.
For long-range contextual feature aggregation, these sparse fusion features are further fed into the GSSM to obtain global gated features.
The RLFB consists of three stacked convolutional blocks followed by ReLU layers, which handle local feature extraction and connect to the upsampling reconstruction stage.
To achieve accurate reconstruction, we repeat this process to extract global cross-modal contextual aggregation features. These are ultimately fed into the RLFB and added via skip connections to produce the final restored image.
Our efficient visual state space model adopts a UNet-based hierarchical encoder-decoder framework. This structure effectively fuses the cross-modal features of events and images while exploring their long-range contextual relationships to output the reconstructed image $I_{restored}$.

The overall framework is trained by minimizing the following loss function:
\begin{equation}
	\mathcal{L} = \left\|I_{restored} - I_{gt}\right\|_{1}
\end{equation}
where $\|\cdot\|_{1}$ indicates the $L_1$ mean absolute error norm and $I_{gt}$ denotes the ground-truth image.
\subsection{Top-k Sparse Attention Module}
Recently, Transformers have seen widespread application in vision tasks. By computing self-attention globally across all tokens, they greatly improve reconstruction networks, but this comes at the cost of significant computational complexity.
To mitigate this limitation, \cite{zhao2019explicit} introduced a top-k selection mechanism for self-attention in NLP tasks to achieve sparse attention and reduce computational costs. Furthermore, \cite{wang2022nformer, knn, chen2023learning} designed KNN-based self-attention mechanisms in spatial dimensions for vision tasks.
Motivated by these advancements, we develop a dynamic top-k selection operation within the state space model. This operation takes advantage of the spatial sparsity of events to selectively fuse complementary image and event features.
Previous event-guided image reconstruction works~\cite{Learning,eventdriven} typically adopt simple multiplication or concatenation of feature maps to represent and fuse auxiliary event information.
However, these naive methods are inefficient and ignore the local sparsity of event features, thereby introducing additional noise and computational overhead~\cite{evlight, sun2025motion}.
To effectively learn cross-modal correspondences, we propose the TSAM. As shown in Figure~\ref{fig:structure}(b), it facilitates interaction between the two modalities under the guidance of top-k selection attention, adaptively selecting receptive fields for different spatial patches.
Our TSAM takes as input queries $\mathbf{Q}_I$ from the image features, along with keys $\mathbf{K}_E$ and values $\mathbf{V}_E$ from the event features, processed using $3 \times 3$ depth-wise convolutional layers and normalized $1 \times 1$ convolutions. 
Next, we calculate the cosine similarity~\cite{choi2024reciprocal} of pixel pairs between the image query and the event key, followed by a top-k selection.
This dynamic selection process shifts the attention from dense to sparse, computed as:
\begin{equation}
    TSAM(\mathbf{Q}_I, \mathbf{K}_E, \mathbf{V}_E) = \tau_{k}\left(\frac{\mathbf{Q}_I^{\mathrm{T}}\mathbf{K}_E}{\sqrt{d_{k}}}\right)\mathbf{V}_E
\label{eq:tsam}
\end{equation}
where $\tau_{k}(\cdot)$ denotes the proposed top-k selection operation, and $d_{k}$ represents the hidden layer dimension. 
We apply this sparse attention across spatial rather than channel dimensions to minimize memory complexity.

\begin{figure}[t!]
    \centering
    \includegraphics[width=\linewidth]{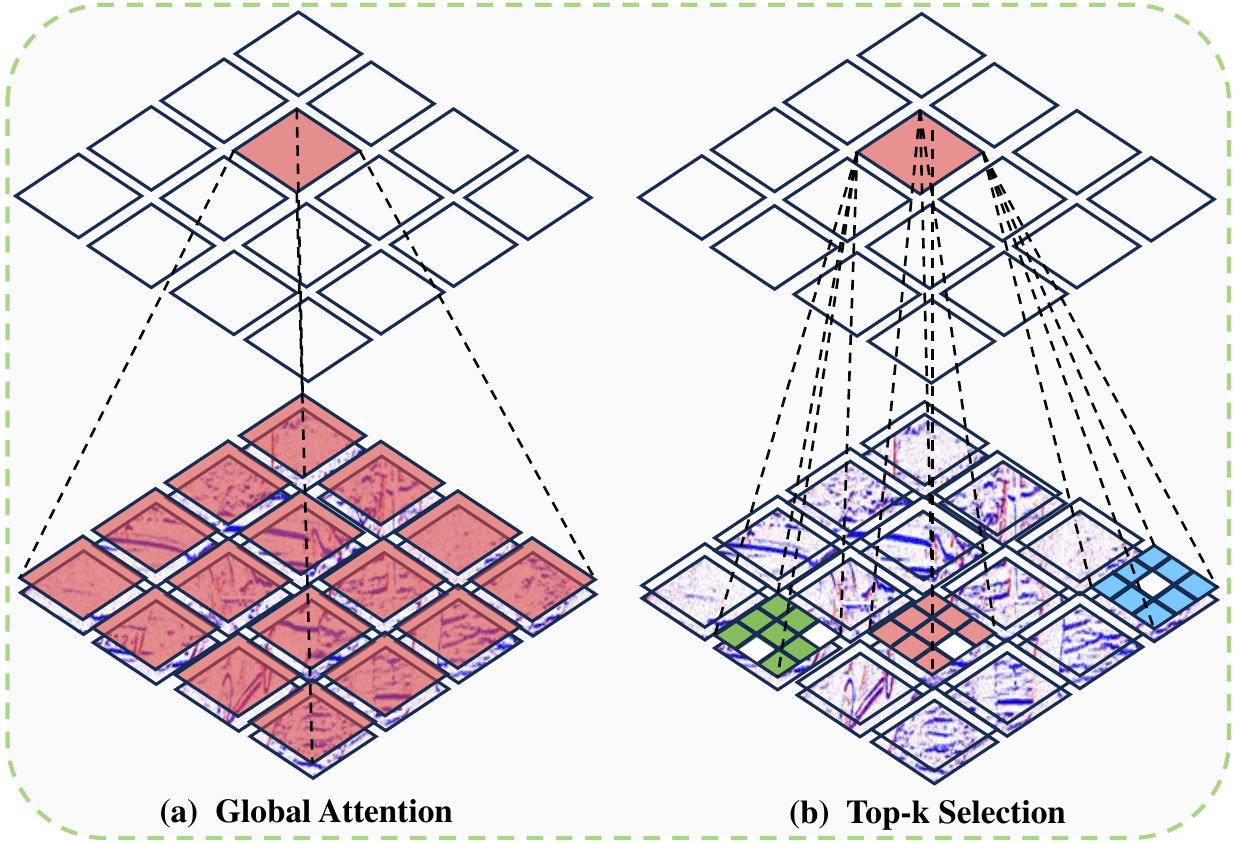}
    \caption{
    Illustration of our top-k selection sparse attention and vision transformer with global attention mechanisms.
    }
\vspace{-0.4cm}
\label{fig:top-k}
\end{figure}

As illustrated in Figure~\ref{fig:top-k}, standard global attention mechanisms in Transformers query and aggregate all patches of the event spatial features (i.e., all red blocks in (a)).

In contrast, our adaptive top-k selection operation aggregates only the top $k$ most similar patches (e.g., $k=3$, represented by the red, green, and blue blocks in (b)). This retains the most critical complementary pixels while discarding uninformative ones.

Specifically, our method adaptively calculates pixel-based contribution scores on the transposed attention matrix $M \in \mathbb{R}^{HW \times HW}$, where $k$ serves as an adjustable parameter dynamically controlling the sparsity level.

Thus, only the top-k values within the interval are normalized from each patch of attention matrix $M$ for softmax computation.  
 
For elements with scores lower than the top-k threshold, we use a scatter function to set their probabilities to zero at specified indices, defined as follows:

\begin{equation}  
    \tau_{k}(M)_{ij} =   
    \begin{cases}  
    M_{ij} & \text{if } M_{ij} \in \textit{top-k}(j) \\
    0 & \textit{otherwise}  
    \end{cases}  
\end{equation}

Although calculating the $M$ matrix involves an $HW \times HW$ multiplication, the top-k selection retains only the sparse $k$ self-attention values by masking each query.

Furthermore, we concatenate all multi-head attention outputs with a small $k$ value and apply a linear projection.

Finally, we execute matrix multiplication with the value matrices $\mathbf{V}$ using sparse matrix multiplication, significantly reducing both computational load and memory usage.

Our TSAM selectively interacts cross-modal features to aggregate complementary event information into fused features, which are then passed to the subsequent GSSM module for efficient global correlation aggregation.

\begin{figure}[t!]
    \centering
    \includegraphics[width=\linewidth]{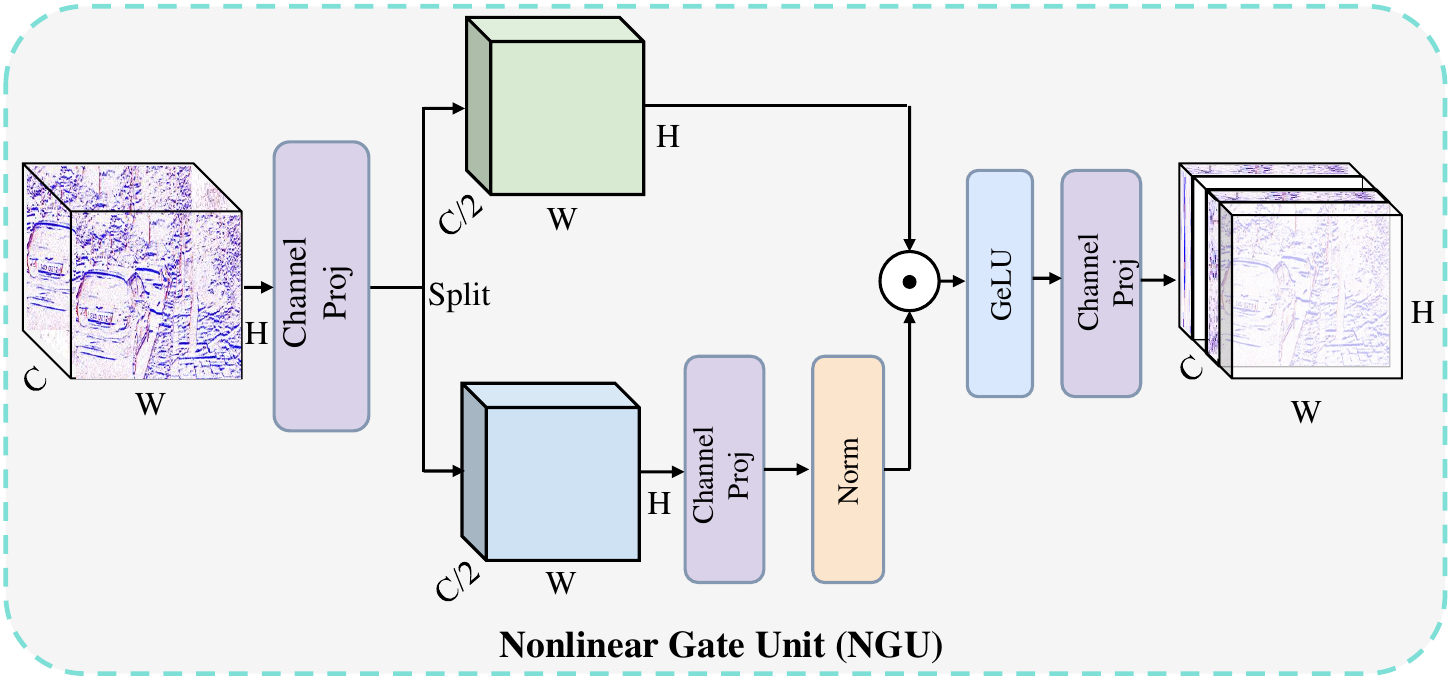}
    \caption{Illustration of the proposed nonlinear gated unit.
    }
\vspace{-0.4cm}
\label{fig:gmlp}
\end{figure}

\subsection{Gated State Space Module}
To efficiently reconstruct high-resolution images, our GSSM is designed to learn a nonlinear mapping of continuous-time events and reduce computational costs along the channel dimension, as depicted in Figure~\ref{fig:structure}(c).
The GSSM integrates sparse fusion features in the underlying channel domain by incorporating the vanilla VSSM~\cite{Vmamba}, capturing essential global contextual information while maintaining high efficiency.
The input fusion feature $\mathbf{X} \in \mathbb{R}^{H \times W \times C}$ passes through three sequential stages.  
In the first stage, we feed the feature into a multi-scale enhancement block with varying convolution sizes to learn multi-scale correlations across different spatial event densities.
Rich multi-scale event representations can effectively capture both fine-grained textures and coarse-grained edge information in the event space~\cite{yang2023event, evlight, lu2024event}.
In the second stage, the multi-scale feature channel is processed by a deep $1 \times 1$ convolution, Layer Normalization (LN), and the Visual SSM layer to extract long-range global features.  
The visual SSM~\cite{SSM} is inspired by continuous linear time-invariant systems, which map a 1D sequence $x(t)$ through an implicit latent state $h(t)$ to output $y(t)$, defined as:

\begin{table*}[t!]
\centering
\resizebox{\textwidth}{!}
{
\begin{tabular}{ccc|ccc|ccc}
\toprule
HDR      & \multicolumn{2}{c|}{SDSD}         &  Deblurring  & \multicolumn{2}{c|}{GoPro}        & Deraining     & \multicolumn{2}{c}{Adobe240}     \\
Methods        & PSNR ↑         & SSIM ↑          & Methods    & PSNR ↑         & SSIM ↑          & Methods      & PSNR ↑         & SSIM ↑          \\
\midrule
esL-Net       & 23.65          & 0.8069          & MPRNet     & 32.66          & 0.9594          & WGWS-Net    & 33.53          & 0.9014          \\
RetinexFormer & 23.76          & 0.8066          & Restormer  & 32.92          & 0.9610           & Histoformer & 30.89          & 0.8709          
\\
Uformer       & 23.91          & 0.8037          & NAFNet     & 33.69          & 0.9672           & FADformer   & 33.41          & 0.8970           \\
Evlight       & 23.93          & 0.7752          & EFNet      & 35.46          & 0.9720          & MPRNet      & 33.79          & 0.8986          \\
EmambaIR      & \textbf{24.15} & \textbf{0.8164} & EmambaIR     & \textbf{35.74} & \textbf{0.9735} & EmambaIR    & \textbf{34.63} & \textbf{0.9027} \\
\bottomrule
\end{tabular}
}
\vspace{-0.2cm}
\caption{
    Quantitative comparison results of our method and other state-of-the-art methods on three reconstruction tasks. 
    }
\vspace{-0.3cm}
\label{table:results}
\end{table*}
\begin{equation}
\begin{aligned}
    h'(t) &= \mathbf{A}h(t) + \mathbf{B}x(t) \\ 
    y(t) &= \mathbf{C}h(t) + \mathbf{D}x(t)
\end{aligned}
\end{equation}
where $t$ denotes the state size, and $\mathbf{A} \in \mathbb{R}^{N\times N}$, $\mathbf{B} \in \mathbb{R}^{N\times 1}$, and $\mathbf{C} \in \mathbb{R}^{1\times N}$ are parameters for state size $N$. $\mathbf{D} \in \mathbb{R}^{1}$ represents the skip connection.
Afterward, the global fusion features are passed to our proposed nonlinear gated unit.

\noindent \textbf{Nonlinear Gated Unit.}
Gated linear units are widely utilized in advanced image restoration algorithms~\cite{zamir2022restormer, liu2021pay, nafnet} and can be formulated as:
\begin{equation}
    Gate(\mathbf{X}, f, g, \sigma) = f(\mathbf{X}) \odot g(\mathbf{X})
\end{equation}
where $f$ and $g$ denote linear transformations, and $\odot$ indicates element-wise multiplication.
Due to variable exposure times, events in a continuous stream occur irregularly, causing significant fluctuations in time intervals. 
This temporal uncertainty makes it difficult for purely linear mappings to accurately capture the relationships between events.
Furthermore, real-world scenes are highly dynamic (e.g., changing lighting conditions or object movements), which directly affects event generation. 
Nonlinear mappings can better adapt to these dynamic changes, thereby improving the robustness and accuracy of the model.
Based on this, we integrate a GeLU nonlinear activation function into the channel-dimension gate unit to capture global information efficiently (see Figure~\ref{fig:gmlp}).
We first divide the feature map into two parts along the channel dimension using channel projection and calculate the channel attention as follows:
\begin{equation}
    NGU(\mathbf{X}, f, g, \sigma) = f(\mathbf{X}) \odot Norm(\sigma(g(\mathbf{X})))
\label{eqn:channel-attention}
\end{equation} 
where $Norm(\cdot)$ denotes the global average normalization operation, which empirically improves model stability and aggregates spatial information into channels. $\sigma$ denotes the nonlinear function.
In summary, by leveraging the temporal continuity of events, we replace high-dimensional matrix multiplication with a simplified channel weighting mechanism via channel gating and nonlinear mapping. This achieves high-resolution reconstruction without compromising performance, demonstrating both the simplicity and effectiveness of our framework.

\begin{table}[h]
\centering
\resizebox{\linewidth}{!}
{
\begin{tabular}{cccccc}
\toprule
           & Event      & PSNR↑ & \#Params(M) & \#FLOPs(G) & Time(ms) \\
\midrule
HINet      & \ding{55}   & 32.57 & 38.67       & 171        & 10.32                \\
MPRNet     & \ding{55}   & 32.56 & 20.13       & 1707       & 292.9               \\
Restormer  & \ding{55}   & 33.39 & 26.09       & 141        & 8.21                \\
REDNet     & \ding{51}   & 33.98 & 9.76        & 160        & 9.56                \\
EFNet       & \ding{51}   & \textcolor{blue}{34.59} & \textcolor{blue}{8.97}        & \textcolor{blue}{107}        & \textcolor{blue}{7.06}                \\
REFID      & \ding{51}   & 34.12 & 88.96        & 209       & 16.08              \\
EmambaIR & \ding{51} & \textcolor{red}{34.96} & \textcolor{red}{6.25}        & \textcolor{red}{86}         & \textcolor{red}{5.87}               \\
\bottomrule
\end{tabular}
}
\caption{
Comparisons of computational cost. The optimal and suboptimal results are highlighted in \textcolor{red}{red} and \textcolor{blue}{blue}.
}
\vspace{-0.6cm}
\label{tab:result2}
\end{table}

\section{Experiments and Analysis}
\subsection{Experiment Settings}
\noindent\textbf{Datasets and Metrics.}
We evaluate our method on three tasks: image motion deblurring, image deraining, and image High Dynamic Range (HDR) reconstruction. These tasks benefit significantly from the high temporal resolution of motion information and the high dynamic range imaging capabilities provided by events.
For the image motion deblurring task, we select the widely used GoPro dataset~\cite{GoPro}, which contains 3214 pairs of blurry and sharp images with a resolution of $1280 \times 720$. We utilize 2103 image pairs for model training and the remaining 1111 pairs for testing. 
To further validate the generalization of our method in real-world scenes, we evaluate on the real-world event-guided image deblurring H2D dataset~\cite{yulearning}. This dataset consists of events captured in real scenarios without an event simulator, providing 603 pairs of real-world data for testing.
For the image deraining task, we adopt the Adobe240~\cite{adobe240} dataset, which consists of 120 video sequences recorded at 240 fps with a resolution of $1280 \times 720$.
We select 50 suitable scene sequences for training and 10 for testing, utilizing the commercial software Adobe Photoshop Lightroom to generate simulated rain streaks.
For the image HDR reconstruction task, we choose the SDSD~\cite{sdsd} dataset, which contains paired real-world sequences of low and high dynamic scenes at $1920 \times 1080$ resolution. We use 125 sequence pairs for training and 25 pairs for testing.
Additionally, we use the open-source event simulator ESIM~\cite{ESIM} to generate noisy event streams based on its default noise model.
For evaluation, we employ Peak Signal-to-Noise Ratio (PSNR) and the Structural Similarity Index Measure (SSIM) as our primary metrics.
\begin{figure*}[t!]
    \centering
    \includegraphics[width=\linewidth]{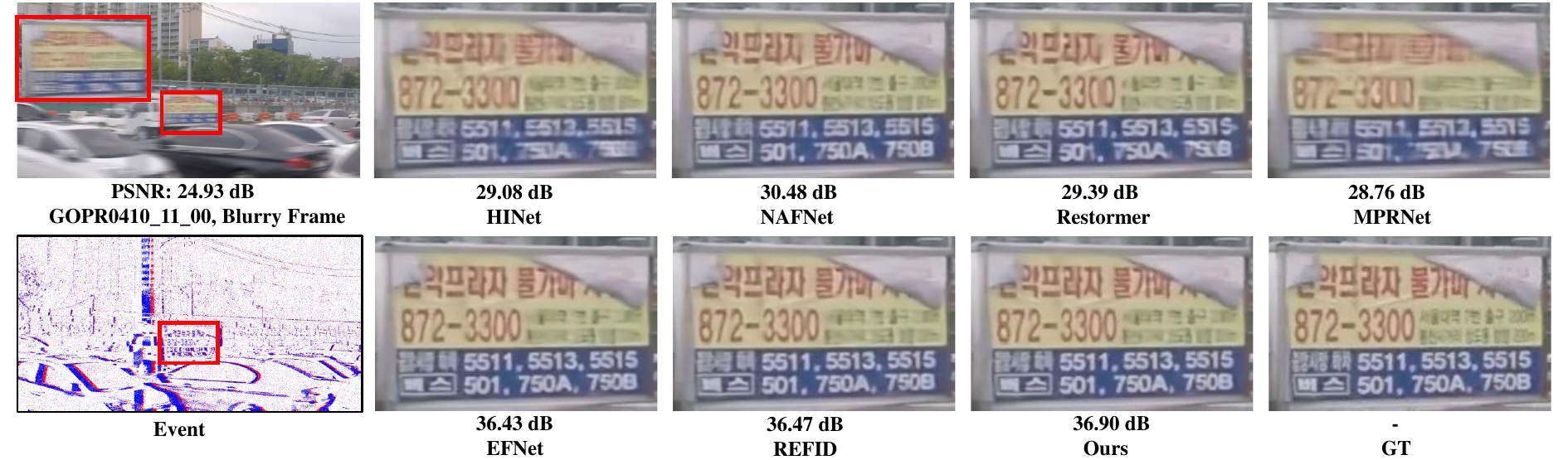}
    \vspace{-0.4cm}
    \caption{
    Qualitative comparison results of different image deblurring methods on the GoPro dataset.
    }
   \vspace{-0.2cm}
\label{fig:deblur_res}
\end{figure*}
\noindent\textbf{Implementation Details}.
All models are implemented in PyTorch and trained on a single NVIDIA GeForce RTX 3090 GPU.
For data augmentation, we employ a range of techniques to improve the model's robustness, including horizontal and vertical flipping, the addition of random noise, and the simulation of hot pixels in event voxels, as described in~\cite{cameras}. These augmentations enhance the model's ability to generalize across diverse scenarios.
The training process is optimized using the Adam optimizer~\cite{Adam} with an initial learning rate of $2 \times 10^{-4}$. 
We apply a cosine annealing schedule to gradually reduce the learning rate to a minimum of $10^{-7}$ over 200,000 iterations, ensuring stable convergence.
Training is performed on $256 \times 256$ crop patches extracted from the full-resolution training data pairs. 
During testing, we evaluate the methods on full-resolution images to validate the model's effectiveness and generalization capabilities across different scenes.
For each reconstruction task, the compared baseline methods include both single-image and event-guided reconstruction approaches.
Note that all compared methods are retrained using the same training data and strategy as our method to ensure a fair comparison.
Remarkably, our model achieves an average training speed that is 20\% faster than the compared methods, while also delivering superior performance.
%
%

%
\begin{figure}[t!]
   \vspace{-0.3cm}
   \centering
    \includegraphics[width=\linewidth]{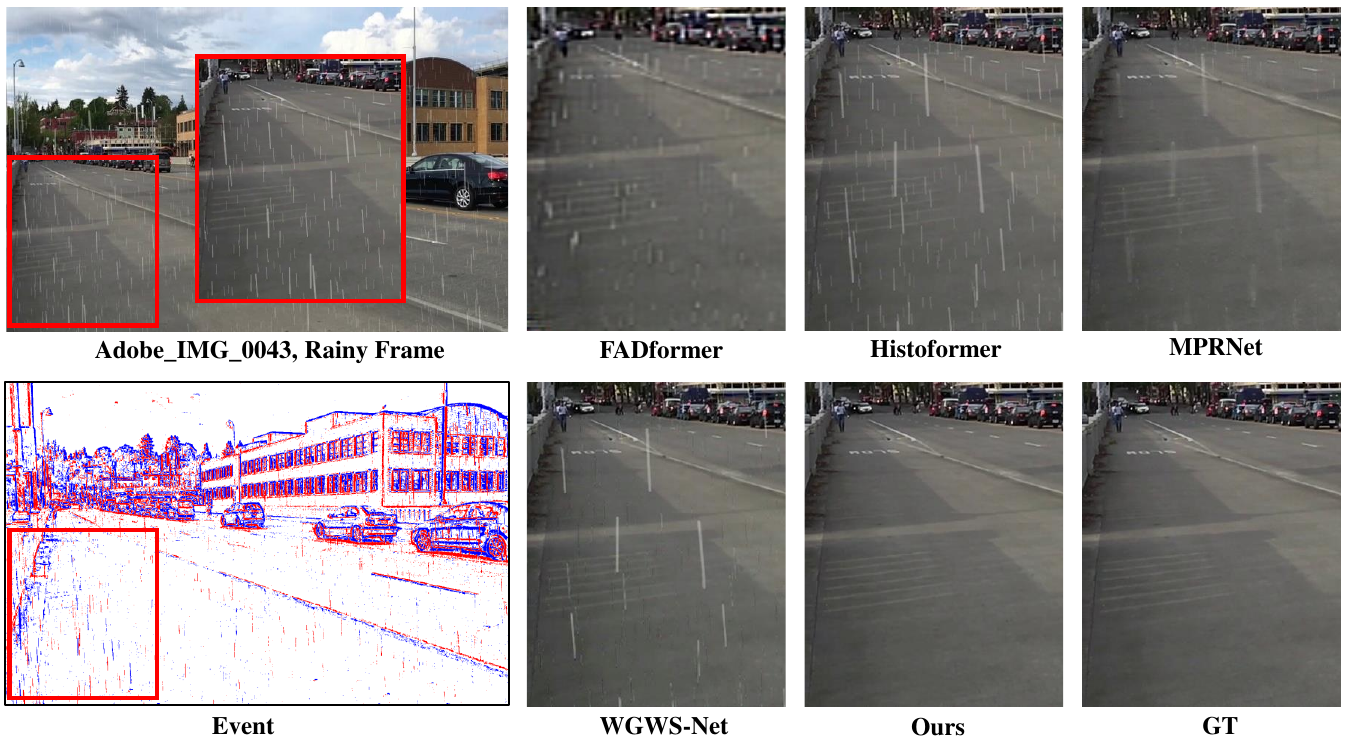}
    \caption{
    Qualitative comparison results of different image deraining methods on the Adobe240 dataset.
    }
   \vspace{-0.4cm}
\label{fig:derain_res}
\end{figure}
\subsection{Comparison Results}
\noindent\textbf{Quantitative Results.}
In Table~\ref{table:results}, we present the quantitative comparison results for the HDR, deblurring, and deraining tasks.
Specifically, for the image HDR task, two image-only HDR methods (RetinexFormer~\cite{cai2023retinexformer} and Uformer~\cite{wang2022uformer}) and two event-guided HDR methods (esL-Net~\cite{eslnet} and Evlight~\cite{evlight}) are selected for comparison.
Table~\ref{table:results} shows that \mname{} outperforms the previously best method, Evlight, by an average of 0.22 dB in PSNR, highlighting the effectiveness of our SSM-based framework.
%
For the motion deblurring task, our approach is compared to advanced image-based and event-guided deblurring methods, including HINet~\cite{chen2021hinet}, NAFNet~\cite{nafnet}, Restormer~\cite{zamir2022restormer}, MPRNet~\cite{MPRNet}, EFNet~\cite{efnet}, and REFID~\cite{refid}.
Our method achieves the highest PSNR and SSIM values among all evaluated approaches. Compared to the state-of-the-art EFNet, our \mname{} achieves a significant improvement of 0.28 dB in PSNR and 0.015 in SSIM. As shown in Table~\ref{tab:result2}, it accomplishes this while maintaining a low parameter count of 6.25M and a computational cost of 86G.
In addition, we compare four image deraining methods: WGWS-Net~\cite{wgws-net}, Histoformer~\cite{Histoformer}, FADformer~\cite{FADformer}, and MPRNet~\cite{MPRNet}.
To the best of our knowledge, our \mname{} is the first event-guided image deraining algorithm.
It can be observed that our method outperforms the best baseline, MPRNet, with an average PSNR improvement of 0.84 dB. This substantial gain directly benefits from the rain streak movement information provided by the event stream.
These results demonstrate that our SSM-based architecture enables the efficient cross-modal fusion and robust utilization of event information for reconstruction, benefiting directly from accurate pixel-level sparse top-k attention and the long-range modeling capability of the gated state space module.

\begin{figure}[t!]
   \vspace{-0.3cm}
   \centering
    \includegraphics[width=\linewidth]{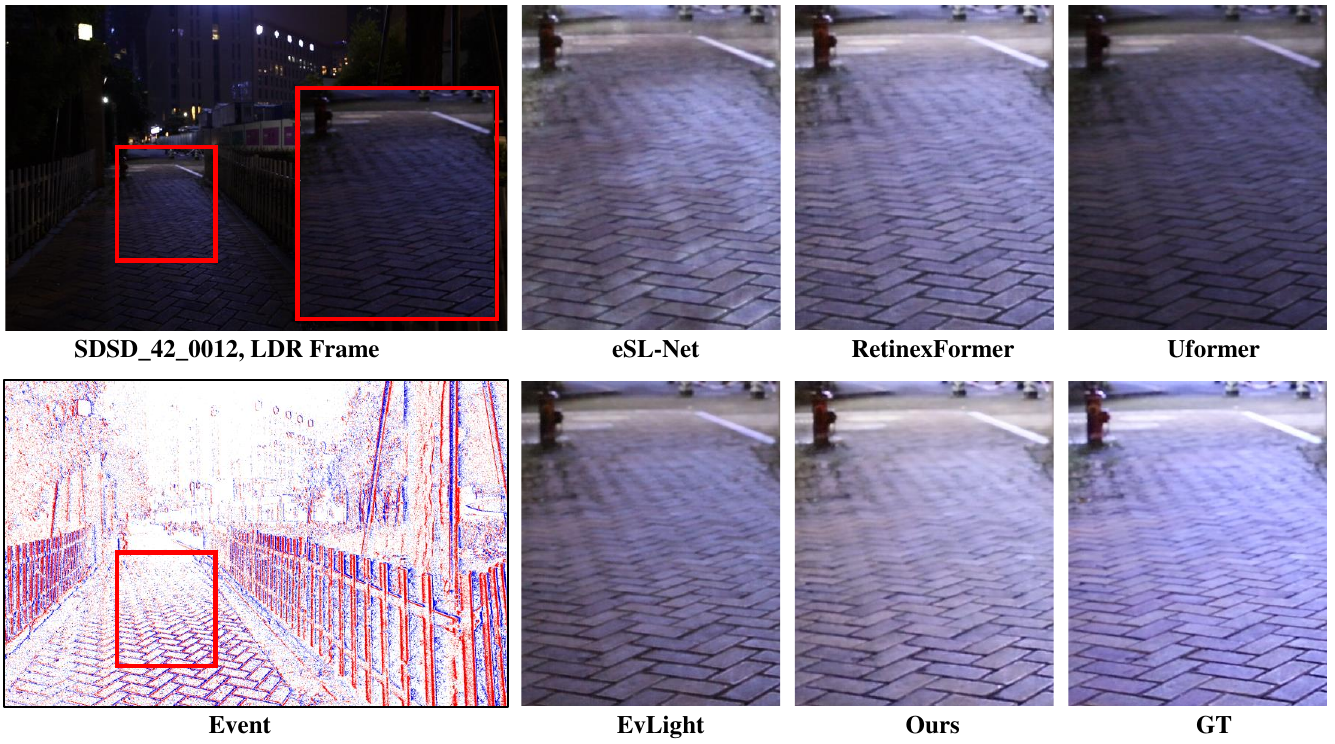}
    \caption{
    Qualitative comparisons of different image HDR methods on the SDSD dataset.
    }
    \vspace{-0.4cm}
\label{fig:hdr_res}
\end{figure}
\noindent\textbf{Qualitative Results.}
Figure~\ref{fig:deblur_res} presents a visual comparison of image deblurring results on the GoPro dataset. Our model effectively removes motion blur and produces sharper, more detailed images. 
As indicated by the accompanying scores, \mname{} visually and quantitatively outperforms the best-performing REFID by 0.43 dB on these samples.
Figure~\ref{fig:derain_res} shows qualitative comparison results on the Adobe240 deraining dataset. 
It is evident that most baseline methods fail to effectively remove dense rain streaks and suffer from noticeable visual artifacts. 
In contrast, \mname{} completely removes the rain streaks and preserves fine background details. By leveraging motion cues from the event stream, our model achieves a clean separation of the moving rain streaks without introducing unwanted artifacts.
Finally, in Figure~\ref{fig:hdr_res}, our method effectively restores underexposed images to reveal intricate structural details; for instance, the individual floor tiles are distinctly resolved.
This is primarily because \mname{} leverages the high dynamic range edge information from the event stream, facilitating a more complete structural recovery and high-fidelity texture restoration compared to other methods.

\begin{table}[h]
\centering
\resizebox{\linewidth}{!}
{
\begin{tabular}{ccccc}
\toprule
Model & \textbf{TSAM} (Space Attention)      & \textbf{GSSM} (Mamba block)       & PSNR↑          & SSIM↑           \\
\midrule
S1    & \ding{55}          & \ding{55}           & 30.23          & 0.9333          \\
S2    & \ding{51}          & \ding{51}           & \textcolor{red}{35.74} & \textcolor{red}{0.9735} \\
S3    & Restormer   & \ding{51}           & 34.91          & 0.9684          \\
S4    & SwinIR    & \ding{51}           & 35.31          & 0.9724          \\
S5    & \ding{51}          & MambaIR      & 35.27          & 0.9718          \\
S6    & \ding{51}          & Freqmamba   & 35.26          & 0.9713          \\
S7    & \ding{51}          & Wave-Mamba & \textcolor{blue}{35.48} & \textcolor{blue}{0.9726}  \\
\bottomrule
\end{tabular}
}
\caption{
Detailed performance comparisons between the proposed modules and their architectural variants.
}
\vspace{-0.2cm}
\label{table:ablation}
\end{table}
\begin{figure}[t!]
   \centering
    \includegraphics[width=\linewidth]{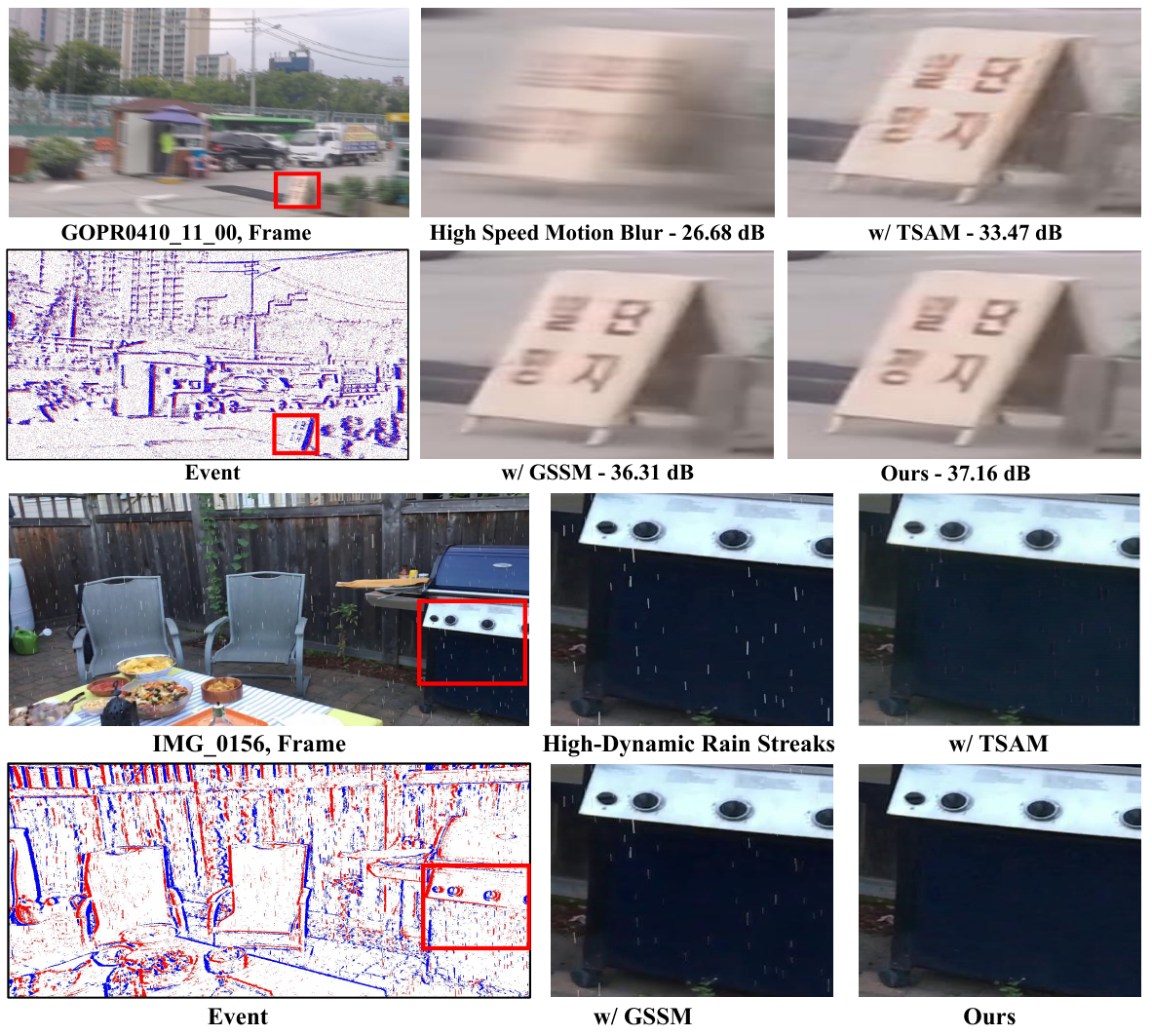}
    \caption{
    Ablation studies of the proposed TSAM and GSSM under challenging scenes with high-speed motion blur and high-dynamic rain streaks.
    }
    \vspace{-0.3cm}
\label{fig:ab1}
\end{figure}
\subsection{Ablation Studies}
\noindent\textbf{Effectiveness of TSAM and GSSM.}
To verify the effectiveness of the proposed TSAM and GSSM in our EmambaIR, we conduct ablation studies to analyze the impact of each module on overall performance.
In Table~\ref{table:ablation}, we evaluate seven model configurations (S1 - S7), including architectural variants where our modules are replaced by existing advanced blocks (e.g., SwinIR~\cite{liang2021swinir} and MambaIR~\cite{guo2025mambair}).
The S1 baseline model stacks only residual local feature blocks without using any SSM. 
Our full model (S2), which incorporates both TSAM and GSSM, outperforms this baseline by over 5.5 dB in PSNR and surpasses all other variants (S3–S7), clearly demonstrating the effectiveness of our design.
Specifically, compared to the SwinIR variant (S4) which relies on dense attention, our full model exhibits an increase of 0.43 dB in PSNR. Furthermore, when compared to the MambaIR variant (S5) based on the vision Mamba mechanism, it achieves an improvement of 0.47 dB.
We also conduct qualitative experiments to evaluate our TSAM and GSSM under extreme conditions, as shown in Figure~\ref{fig:ab1}.
Both modules demonstrate strong performance in high-speed motion deblurring. 
In low dynamic range scenes where the background image is nearly invisible, the model equipped with TSAM effectively removes all rain streaks, whereas the model utilizing only GSSM leaves some residual artifacts. 
This highlights the importance of TSAM's spatially selective aggregation, which plays a critical role in enhancing reconstruction quality. 
Overall, these results confirm that TSAM and GSSM complement each other perfectly, fully exploiting event-based features to restore structures and details in challenging high-speed and high-dynamic range scenarios.

\begin{figure}[t!]
   \centering
    \includegraphics[width=0.95\linewidth]{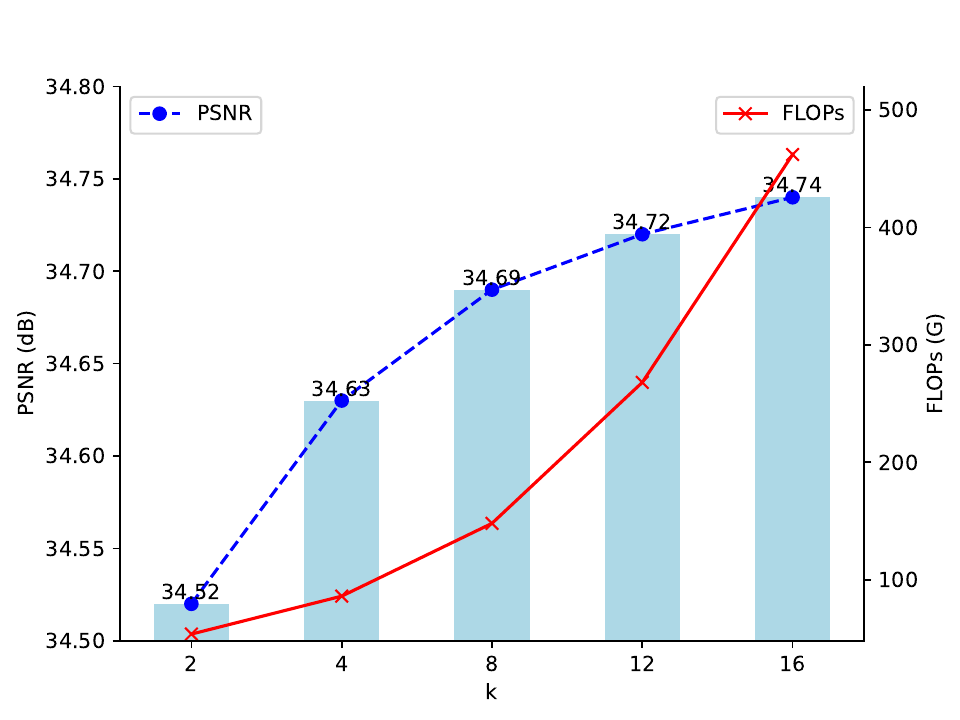}
    \vspace{-0.3cm}
    \caption{
    Ablation studies on the impact of varying $k$ values in TSAM.
    }
\vspace{-0.3cm}
\label{fig:ab-k}
\end{figure}

\noindent\textbf{Effectiveness of $k$ in TSAM.}
The hyperparameter $k$ determines the maximum number of patches involved in the local sparse pixel-level attention via the top-k selection process. 
To validate the impact of $k$, we present the ablation results in Figure~\ref{fig:ab-k}.
It is worth noting that adjusting $k$ does not alter the number of network parameters; it only influences memory usage and computational complexity. 
As expected, increasing the value of $k$ yields higher PSNR performance but incurs greater computational costs, which confirms the auxiliary benefit of event sparse features on image reconstruction.
Although setting $k = 16$ achieves the highest absolute performance, we set $k = 4$ as our default configuration because it offers highly comparable performance while significantly reducing computational complexity.
Ultimately, this allows the performance and efficiency of our \mname{} to be flexibly balanced according to the specific hardware or latency constraints of different real-world applications.

\section{Conclusion}
In this work, inspired by the spatial sparsity and temporal continuity of event streams, we proposed an efficient visual state-space model for event-guided image reconstruction, dubbed EmambaIR. Our framework utilizes TSAM and GSSM to achieve high-quality cross-modal fusion reconstruction.
Specifically, TSAM dynamically controls feature sparsity and selectively fuses cross-modal information under the guidance of top-k sparse attention.
Subsequently, GSSM employs a nonlinear gated unit to enhance the temporal representation capabilities of vanilla linear-complexity SSMs, thereby drastically reducing the computational overhead typically associated with high-resolution reconstruction.
Extensive experiments demonstrate that our EmambaIR outperforms state-of-the-art methods across multiple tasks while maintaining significant advantages in both memory consumption and computational cost.
While this work primarily focuses on image reconstruction tasks, extending this event-guided efficient architecture to video reconstruction tasks—such as video deblurring, video HDR, and video frame interpolation—remains a promising direction for future research.

\bibliography{references}
\bibliographystyle{plain}

\end{document}